\documentclass[letterpaper]{article} % DO NOT CHANGE THIS
\usepackage{aaai24}  % DO NOT CHANGE THIS
\usepackage{times}  % DO NOT CHANGE THIS
\usepackage{helvet}  % DO NOT CHANGE THIS
\usepackage{courier}  % DO NOT CHANGE THIS
\usepackage[hyphens]{url}  % DO NOT CHANGE THIS
\usepackage{graphicx} % DO NOT CHANGE THIS
\usepackage{natbib}  % DO NOT CHANGE THIS AND DO NOT ADD ANY OPTIONS TO IT
\usepackage{caption} % DO NOT CHANGE THIS AND DO NOT ADD ANY OPTIONS TO IT
\usepackage{algorithm}
\usepackage{algorithmic}

\usepackage{booktabs}
\usepackage{amsmath}
\usepackage{amsfonts,amssymb}
\usepackage{subcaption}
\newtheorem{myDef}{Definition}

\usepackage{newfloat}
\usepackage{listings}
\DeclareCaptionStyle{ruled}{labelfont=normalfont,labelsep=colon,strut=off} % DO NOT CHANGE THIS
\floatstyle{ruled}
\newfloat{listing}{tb}{lst}{}
\floatname{listing}{Listing}
\nocopyright
\title{Hyper-Relational Knowledge Graph Neural Network for Next POI Recommendation}
\author{
    Jixiao Zhang\textsuperscript{\rm 1}, Yongkang Li\textsuperscript{\rm 2}, Ruotong Zou\textsuperscript{\rm 1}, Jingyuan Zhang\textsuperscript{\rm 1}, Zipei Fan\textsuperscript{\rm 3}, Xuan Song\textsuperscript{\rm 1}
}
\affiliations{
    \textsuperscript{\rm 1} Southern University of Science and Technology\\
    \textsuperscript{\rm 2} University of Amsterdam 
    \textsuperscript{\rm 3} University of Tokyo 
}

\usepackage{bibentry}
\begin{document}

\maketitle

\begin{abstract}
With the advancement of mobile technology, Point of Interest (POI) recommendation systems in Location-based Social Networks (LBSN) have brought numerous benefits to both users and companies. Many existing works employ Knowledge Graph~(KG) to alleviate the data sparsity issue in LBSN. These approaches primarily focus on modeling the pair-wise relations in LBSN to enrich the semantics and thereby relieve the data sparsity issue. However, existing approaches seldom consider the hyper-relations in LBSN, such as the mobility relation~(a 3-ary relation: user-POI-time). This makes the model hard to exploit the semantics accurately. In addition, prior works overlook the rich structural information inherent in KG, which consists of higher-order relations and can further alleviate the impact of data sparsity.To this end, we propose a Hyper-Relational Knowledge Graph Neural Network~(HKGNN) model. In HKGNN, a Hyper-Relational Knowledge Graph~(HKG) that models the LBSN data is constructed to maintain and exploit the rich semantics of hyper-relations. Then we proposed a Hypergraph Neural Network to utilize the structural information of HKG in a cohesive way. In addition, a self-attention network is used to leverage sequential information and make personalized recommendations. Furthermore, side information, essential in reducing data sparsity by providing background knowledge of POIs, is not fully utilized in current methods. In light of this, we extended the current dataset with available side information to further lessen the impact of data sparsity. Results of experiments on four real-world LBSN datasets demonstrate the effectiveness of our approach compared to existing state-of-the-art methods.
\end{abstract}

\section{Introduction}
Point of Interest (POI) recommendation systems aim to provide personalized and relevant recommendations of the most likely visited POI for users based on their location, interests, social relationships, and past behaviors within the context of Location-based Social Networks (LBSN). The widespread availability of mobile devices and social networks has made it easier for users to access and share check-ins and POI information, leading to an increased focus on POI recommendations in recent years.\par
 
Previous researchers have proposed numerous methods that model social and check-in relations in LBSN from various perspectives, aiming to facilitate the next POI recommendations.
Early studies employed collaborative filtering techniques, such as Matrix Factorization (MF)~\cite{koren2009collaborative}, to model user preferences, while others utilized Markov Chains (MC) to leverage sequential information. 
Recent studies predominantly utilize Recurrent Neural Networks (RNNs)-based models or attention mechanisms to capture the sequential information in the trajectories of users and incorporate spatial and temporal factors in distinct ways.
Due to the successful application of self-attention in language models, which demonstrates exceptional potential in extracting sequential information, several state-of-the-art models have replaced RNNs with self-attention. For instance, GeoSAN~\cite{lian2020geography} utilizes a geography-aware self-attention network. Furthermore, to capture higher-order relations between POIs and users, recent studies have also employed Graph Neural Networks (GNNs) to take the structural information into account\cite{asgnn,lim2022hierarchical}.

\begin{figure}[t]
\includegraphics[width=\linewidth]{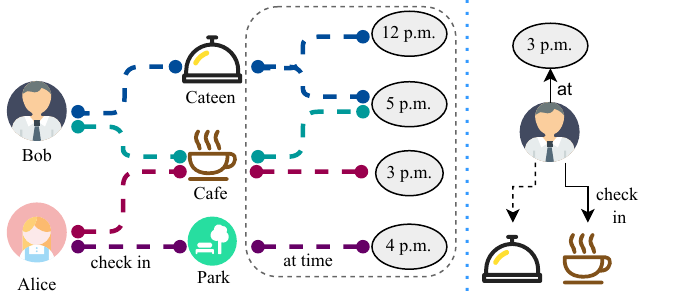}
\caption{Examples of hyper-relations in LBSN}
\label{contribution}
\end{figure}

However, despite the strong capability of these models in modeling user preferences and achieving excellent performance.  The data sparsity in LBSN still remains a major challenge that impedes future advancements.  LBSN exhibits a long-tail distribution, with numerous POIs having limited check-ins. Existing methods face challenges in modeling user preferences for those POIs since they seldom appear in the trajectories of users. Consequently, it is hard to make accurate recommendations based solely on social relations and check-in data.  \par
In recent years, Knowledge Graph (KG) has emerged as a valuable tool for managing side information and offering in-depth semantic information about entities and relations, which have shown significant potential in mitigating the issue of data sparsity in general recommendation systems~\cite{zhang2016collaborative, wang2018dkn, wang2019kgat}.
Several studies have also employed KG as a component of their models in the field of the next POI recommendation \cite{qian2019spatiotemporal, zhang2020kean, chen2022building} to alleviate the data sparsity issue. However, they face the following limitations that remain unsolved: \textbf{1)} Existing approaches prioritize pairwise KG relations, neglecting hyper-relations in LBSN. An example is the mobility relation.  Prior methods treat mobility as user-POI pair as illustrated in Fig.~\ref{contribution}~(leftmost), which failed to capture the time-dependent features. For instance, if we want to predict where Bob will check in at 3 p.m., the model will struggle with whether to recommend the canteen or the cafe~(rightmost). By taking time into consideration(middle), recommending the cafe is apt due to its afternoon popularity, whereas the canteen suits lunch and dinner times. \textbf{2)} Previous works either employ KG only for embedding initialization \cite{guo2020attentional, qian2019spatiotemporal} or treat the next POI recommendation problem as a KG completion task \cite{wang2021spatio}. These approaches primarily focus on utilizing the semantic information in KG while disregarding the valuable structural information. \textbf{3)} Current methods overlook the potential of fully utilizing the available side information in LBSN. \citet{wang2021spatio} have introduced multi-level category relations of POIs. Other side information, such as average price, ratings, etc., remains untapped. Leveraging this additional information can offer valuable insights for modeling user preferences, particularly for the least visited POIs.\par
To overcome the limitations discussed earlier, we propose a novel \underline{H}yper-relational \underline{K}nowledge \underline{G}raph \underline{N}eural \underline{N}etwork (HKGNN) for the next POI recommendation. HKGNN first constructs a Hyper-Relational Knowledge Graph (HKG) by incorporating the mobility pattern relations (i.e., the check-ins), social relations, and side-information relations (side-info relation, for short, is used in some cases). Then, the HKG is directly transformed into a hypergraph,  and a Hyper-Graph Neural Network (HGNN) is applied to capture the higher-order relations by effectively utilizing the structural information of the HKG. In this way, we exploit both the semantics and structural information of the HKG cohesively. 
Furthermore, we proposed Knowledge-aware Self-Attention Encoder(KAAE) and Spatio-Temporal Attention Decoder(STAAD)to exploit the rich sequential information in LBSN. 
Additionally, to enhance the semantics in the HKG, we also extend the current dataset with available POI side information.
In summary, the main contributions of this paper are as follows:

\begin{itemize}
\item  We propose a novel Hyper-relational Knowledge Graph Neural Network~(HKGNN) capable of modeling any hyper-relations in LBSN. To the best of our knowledge, HKGNN is the first work that incorporates the HKG in the next POI recommendation task.
\item We proposed a Hyper-Graph Neural Network~(HGNN) to leverage the structural information of HKG in a cohesive way.
\item We extend current datasets with POI side information, and the experiments on least visited POIs demonstrate the effectiveness in alleviating data sparsity issues.
\item Extensive experiments were conducted on four real-world LBSN datasets and achieved state-of-the-art with significant improvements. The ablation study further highlights the effectiveness of each component of the model.
\end{itemize}

% The remaining sections of this paper are organized as follows. Section 2 provides a review of existing research on the next POI recommendation. In Section 3, we present the definitions related to this study and the problem formulation. Section 4 introduces the construction of the HKG and our proposed HKGNN. Our conducted experiments are presented in Section 5. Finally, Section 6 concludes the paper. 

\section{Related Work}

LBSN offers rich content and is commonly utilized for two main tasks: friend recommendation \cite{friend1, friend3} and POI recommendation\cite{feng2015personalized, poi2}. In contrast to traditional recommendations such as goods or book recommendations, POI recommendation focuses on incorporating spatiotemporal information and social networks from LBSN. There are two main types of POI recommendation tasks: \textit{general} and \textit{next} POI recommendation. The key distinction between next POI recommendation and general POI recommendation lies in their objectives. Next POI recommendation aims to determine the transition probabilities between POIs for each user, whereas general POI recommendation seeks to understand the overall interests of all users. Given a user and their past check-in records, the next-POI recommendation system gives a POI that the user is likely to visit next.\looseness=-1\par
In recent studies, many approaches have involved KG to alleviate the data sparsity issue. Meta-SKR \cite{meta2021} introduced a sequential knowledge graph with a meta-learning module. STKG \cite{wang2021spatio} constructed a spatiotemporal knowledge graph and considered POI recommendation as a KG completion task. STKGRec\cite{chen2022building} further combined STKG with the next POI recommendation model, which can capture both long- and short-term preferences of users.  However, current works seldom consider hyper-relations in LBSN.\par
The structural information in LBSN is crucial for POI recommendation. In SGRec~\cite{ijcai2021p206}, it transforms the check-in sequences into graphs and exploits the POI transition pattern. STP-UDGAT~\cite{lim2020stp} directly builds POI-POI spatial, temporal, and transition graphs based on check-in data. Then it exploits a Graph Attention Network(GAT) to model the structural information.  However, the structural information is seldom considered in KG-based models. GraphFlashback~\cite{rao2022graph} utilizes KG as an embedding initialization method and only extracts a POI transition graph from the KG for capturing the sequential transition patterns. We argue that such methods underutilize the structural information in KG.

\section{Preliminaries}
In this section, we commence by presenting a formal definition of the LBSN data. Subsequently, we define the concepts of HKG. Then, we introduce hypergraph that is central to our proposed approach. Finally, we provide a precise definition of the next POI recommendation task. \par

\textbf{LBSN Data}
Let $U=\{u_1,u_2,\ldots,u_{|U|}\}$ be a set of users, 
$P=\{p_1,p_2,\ldots,p_{|P|}\}$ be a set of POIs. For each user $u_i$, each of its check-ins $c$ is represented by $\left(u_i,p_j, t_k\right)$, which indicates that POI $p_j$ has been visited by user $u_i$ at time $t_k$. And we use $F=\{(u_1, u_i), \dots, (u_j, u_k) \}$ to be a set of friendships in LBSN.
In addition, let $L=\{\ell_1, \ell_2, \dots, \ell_{|P|} \}$ be a set of locations, where $\ell_k$ is the location of POI $p_k$.
Also, let $A = \{A_1, A_2, \dots, A_{|A|} \}$ be the POI side information set. Specifically, $A_i=\{A_{i,1}, A_{i,2}, \ldots, A_{i, \mathit{m}} \}$ represents one specific side information type, where $m$ is the number of possible values.
And each user $u_i$ has a historical check-in trajectory 
$H_{u_i}=\{c_1, c_2, \dots, c_n\}$, where $n$ is the length of trajectory.
And $\mathcal{H}_U=\{H_{u_1},H_{u_2},\ldots,H_{u_{|U|}}\}$ contains all historical check-in records of users. \looseness=-1\par

\begin{myDef}[Hyper-Relational Knowledge Graph]\rm
A Hyper-relational Knowledge Graph~(HKG) is mathematically defined as $\mathcal{G}_{kg} =\left(\mathcal{E}, \mathcal{R},  \mathbf{E}_{kg} \right)$, where $ \mathcal{E} = \{e_1, e_2, \dots, e_{|\mathcal{E}|}\}$ is a finite set of entities, $\mathcal{R} = \{r_1, r_2, \dots, r_{|\mathcal{R}|}\}$ is a finite set of relations and $\mathbf{E}_{kg}= \{r\left(e_1, e_2, \dots, e_{k}\right)| e_1, \dots, e_k \in \mathcal{E}, r \in \mathcal{R},\} $ is a finite set of facts in HKG, specifically, $r\left(e_1, e_2, \dots, e_{k}\right)$ is a fact in the form of a tuple and represents that entities $e_1, e_2, \dots, e_{k}$ have a relation $r$. To facilitate the representation of entities in the HKG, an embedding set $H=\{\vec{h_1}, \vec{h_2}, \dots, \vec{h}_{|\mathcal{E}|}\}$ is introduced, where each element $\vec{h_i}$ corresponds to an entity and is represented as a vector in $\mathbb{R}^d$ space, with $d$ being the embedding dimension. In addition, to leverage the structural information in HKG, we directly transformed the HKG into a hypergraph $\mathcal{G} = \left(\mathcal{V}, \mathbf{E}\right)$.$\mathcal{V}$ represents a set of vertices, and $\mathbf{E}$ is a finite set of hyperedges.
\end{myDef}

\begin{figure}[t]
\includegraphics[width=\linewidth]{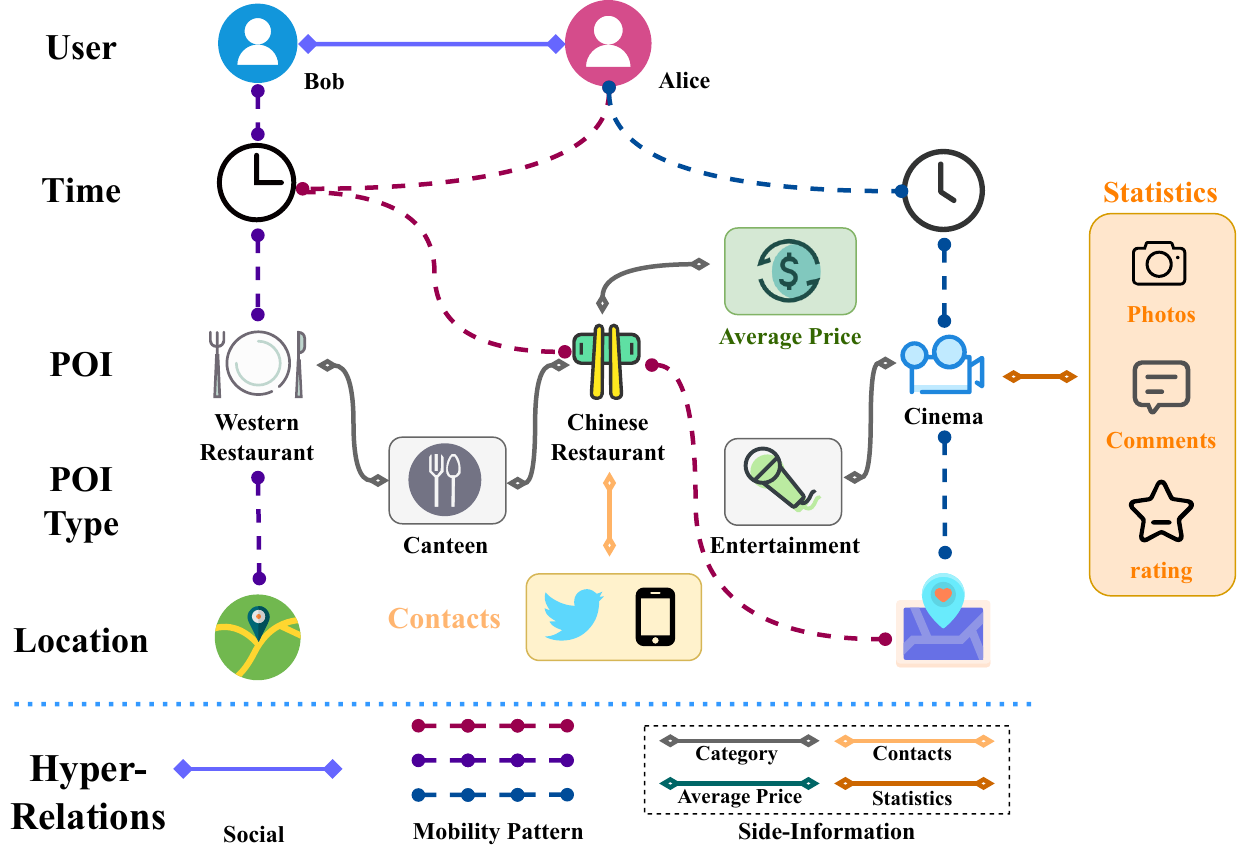}
\caption{A simple illustration of our HKG}
\label{HKG}
\end{figure}

\textbf{Problem Formulation}
Given user $u_i$ with corresponding trajectory $H_{u_i}$ and next check-in time $t_k$, the goal of the next POI recommendation is to predict the most likely visited POI for user $u_i$ at time $t_{k}$.\par

\begin{figure*}[t]
\includegraphics[width=\textwidth]{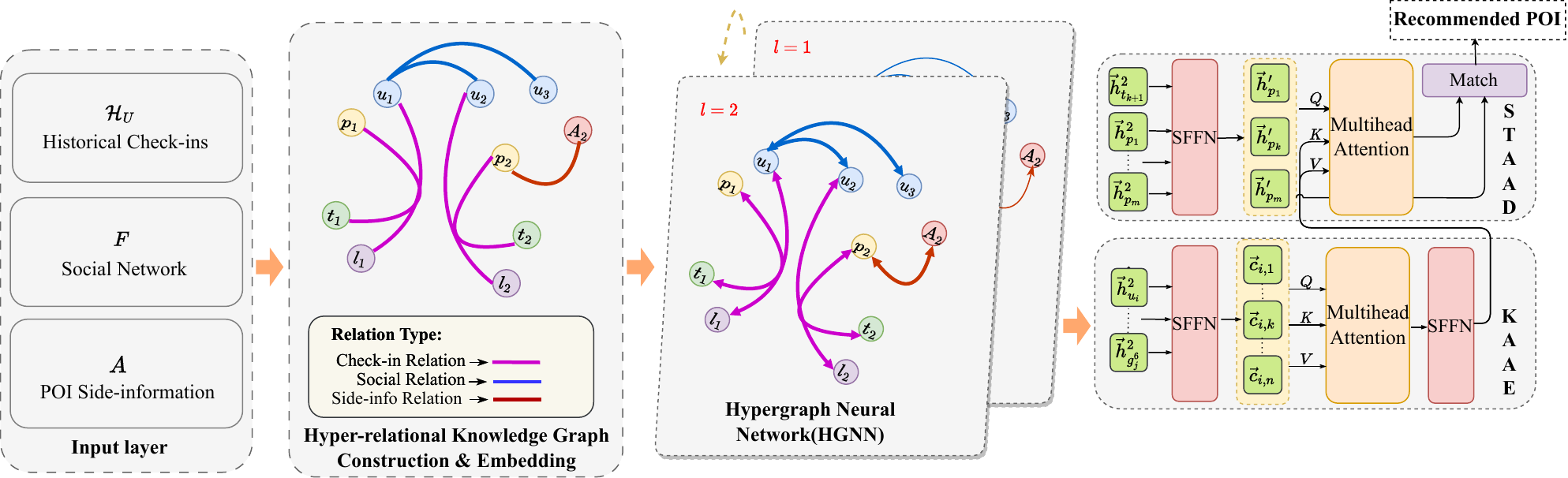}
\caption{The overview of our HKGNN model}
\label{Framework of HKGNN}
\end{figure*}

\section{Methodology}
In this section, we propose our HKGNN model, and the overall framework is illustrated in Figure~\ref{Framework of HKGNN}.  Our algorithm starts with preprocessing of the LBSN data, followed by the construction of an HKG. Then, the HSimpE method is employed to embed the entities from the semantics of the HKG. The HGNN then refines the embeddings of entities by incorporating structural information. Subsequently, the representations of the check-in sequence are obtained by aggregating embeddings of entities in check-ins and then fed into the Knowledge-Aware Self-Attention Encoder (KAAE) and the Spatiotemporal-Aware Attention Decoder (STAAD) to capture the sequential information and model the user preference, respectively. Finally, the recommended POI is given by the matching score of user preference and time-encoded POI embeddings.

\subsection{Hyper-Relational Knowledge Graph Construction\label{hkg construction}}
In this paper, we construct a novel hyper-relational knowledge graph~(HKG) consisting of three types of relations: mobility pattern relation, social relations, and POI side-information relation based on LBSN data. A simple illustration is shown in Figure \ref{HKG}. \par
\textbf{Mobility Pattern Relation} Mobility pattern relation is extracted from the check-ins, which indicates that a user visited a POI at a certain time and location. The mobility pattern relation $r_{mobile}$ is defined as $r_{mobile}\left(u_i, p_j, \ell_j, t_k\right)$.\par 
While keeping the fine-grained time partition is important for capturing the mobility pattern at different times, the data sparsity could be vital since only a small number of check-ins exist in a short period. Therefore, we consider the weekly periodicity of user behaviors mentioned in \cite{li2018next} and partition each day in a week into 48 time slots. Also, to express the time more accurately, we take month and year into account. Then, each time $t_k$ can be expressed as $t_k=\left(t^d_k, t^m_k, t^y_k\right)$, where $t^d_k$ is the time partition in a week and $t^m_k, t^y_k$ is the corresponding month and year.\par
As the locations of POIs composed of longitude and latitude are difficult to utilize directly, we employ geohash, which partition POIs into distinct regions, and close locations are encoded into the same hash value. Furthermore, different lengths of the resulting string can be used to obtain regions of varying sizes. To model the spatial relations of POIs at different distances, we utilize geohash lengths of 4, 5, and 6, resulting in precision levels of 20 km, 2.4 km, and 0.6 km, respectively. These correspond to city-wise, district-wise, and street-wise distance relations, denoted by $g^4, g^5, g^6$. Then, the location $\ell_j$ of POI $p_j$ can be written as $\ell_j = \left(g^4_j, g^5_j, g^6_j\right)$\par
\textbf{Social Relation}
A major component in LBSN is the social network, i.e., friendships of users. Friends may have a similar preference for POIs. E.g., a pair of friends may usually eat at the same restaurant. Therefore, social relations can be a great help when making personalized recommendations and when the user has little check-in data. For a pair of friends $\left(u_i, u_j\right) \in F$, we have $r_{social}\left(u_i, u_j\right)$.\par
\textbf{POI side-information Relation}
We consider five types of POI side-info relations: coarse-grained and fine-grained categories, statistics, average price, and contact methods, denoted by $A_{C_1}, A_{C_2}, A_{S}, A_{AP}, A_{CT}$, respectively.\
We incorporate a two-level category relation to model this with different perspectives, denoted by $r_{C_1}=\left(p_i, A_{C_1, i}\right)$ and $r_{C_2}=\left(p_i, A_{C_2, i_1}, \dots, A_{C_2, i_k}\right)$, where $A_{C_1, i}$ represents a coarse-grained level consisting of categories such as food and entertainment, and $A_{C_2, i_k}$ represents a fine-grained level consisting of more specific categories such as Chinese restaurants and art gallery. And a POI might have multiple second-level categories, e.g., a restaurant is both a Japanese restaurant and a buffet.
We also leverage the ratings and the number of comments, likes, and photos as statistics side-info relation, denoted by $r_{S}=\left(p_i, A_{S, rat_i}, A_{S, \#com_i}, A_{S, \#lik_i}, A_{S, \#pho_i}\right)$, where $\#$ indicates the number of each statistic. Each statistic is divided into six possible values, from smallest to largest. In addition, we have average price side-info relation $r_{AP}=\left(p_i, A_{AP, i}\right)$. And there are four price tiers for the POIs. Finally, contact method side-info relation $r_{CT}=\left(p_i, A_{CT, i_1}, \dots, A_{CT, i_k}\right)$, indicating the $i_k$ contact method of $p_i$, is introduced, since some users may use an online platform like Twitter or Facebook to decide where to go next, while some POIs only have a phone number to contact with or even do not provide any contact method.

\subsection{Hyper-Relational Knowledge Graph Embedding \label{hkg emb}}
The embedding model is an effective method to learn the dense representation of entities from the facts in HKG. The HKG embedding technique employs a specific scoring mechanism to evaluate the credibility of the facts in HKG. Subsequently, the embedding vectors are improved by maximizing the credibility of all facts in HKG and thus can capture the semantics.
In our work, we adopt a widely used HKG embedding method, HSimplE\cite{hsimple}. It holds the opinion that each entity has multiple representations at different positions in the hyper-relation and defines a score function as follows,

\begin{small}
\begin{equation}
\phi\left(r\left(e_1,\dots, e_k\right)\right) \!=\! \odot\left(\vec{h_r}, \operatorname{sft}\left(\vec{h_{e_i}}, \frac{d\cdot\left(i -1\right)}{\alpha}\right),\cdots \right),
\end{equation}
\end{small}
where $\vec{h_r}$ is the embedding of relation $r$, $\vec{h_{e_i}}$ is the embeddings of entities $e_i$. $\operatorname{sft}\left(\vec{h_i}, x\right)$ shift $\vec{h_i}$ to the left by $x$ steps. $\alpha=\max _{r \in \mathcal{R}}\left(|r|\right)$ is the max arity in HKG, and $\odot\left(\cdot\right)$ is a variadic function defined as,
\begin{equation}
\odot\left(\vec{h_1}, \vec{h_2}, \dots, \vec{h_k}\right) = \sum_{i=1}^{d} \vec{h_1}^{\left(i\right)} \vec{h_2}^{\left(i\right)}\dots\vec{h_k}^{\left(i\right)},
\end{equation}
where $\vec{h_j}^{\left(i\right)}$ is the $i$-th element of vector $\vec{h_j}$.\par
Negative sampling is used to train the embedding model. For every sample in HKG, we generate $N|r|$ negative samples by replacing each entity in the relation with randomly picked $N$ entities and let $T_{\text{neg}}$ be a function to generate negative samples as above process, where $N$ is the ratio of negative samples. Then binary cross entropy is used as the objective function,
\begin{equation}
    \mathcal{L}_{KG}=\sum_{x \in \mathbf{E}_{kg}}-\log \left(\frac{e^{\phi\left(x\right)}}{e^{\phi\left(x\right)}+\sum_{x' \in T_{\text {neg }}\left(x\right)} e^{\phi\left(x'\right)}}\right)
\end{equation}

\subsection{Hypergraph Neural Network \label{hgnn}}
To leverage the structural information of HKG in a cohesive way, we incorporate the HGNN to refine the representation of entities. Clique Expansion (CE)\cite{ce} and Star Expansion (SE)\cite{star} are two popular methods to tackle hypergraphs, which convert hypergraphs into pair-wise graphs while keeping most structural information. For simplicity, we use the CE method and then employ GAT\cite{gat2018} to model the higher-order relation.
\subsubsection{HKG transformation and expansion}
In the first stage, we transform the HKG into a hypergraph $\mathcal{G}=\left( \mathcal{V}, \mathbf{E}\right)$, where we take $\forall r\left(e_1,e_2,\dots,e_k\right)\in \mathbf{E}_{kg}$ as a hyperedge $E = \left(e_1, e_2, \dots, e_k\right) \in \mathbf{E}$ and $\forall e_i \in \mathcal{E}$ as vertices $v_i \in V$. A hypergraph can alternatively be represented by its incidence matrix $\mathbf{H}$, where $\mathbf{H}_{vE} = 1$ if $v\in E$, and $\mathbf{H}_{vE} = 0$ otherwise. The CE of a hypergraph is to convert each hyperedge into a clique, i.e., each vertex in a hyperedge is connected to all the other vertices. Hence we can get the CE hypergraph $\mathcal{G}_{ce} = \left(\mathcal{V}, \mathbf{E}'\right)$, where $\mathbf{E}'$ is the expanded pair-wise edges, and the adjacent matrix $\mathbf{A}$ of the CE hypergraph is given by:
\begin{equation}
\mathbf{A} = Sgn\left(\mathbf{H}\cdot \mathbf{H}^T\right),
\end{equation}

where $Sgn\left(\cdot\right)$ is a vectorized sign function that maps each value in the matrix into $1$ if it is greater than $0$ and $0$ otherwise.\looseness=-1
\subsubsection{Message propagation on CE hypergraph}
In the second stage, a 2-layer GAT network is used for message propagation on the CE hypergraph $\mathcal{G}_{ce}$. The output embedding $\vec{h}_{i}^{2}$ of each vertex $v_i$ is obtained by:
\begin{small}
\begin{equation}
\vec{h}_{i}^{l}=\sigma\left(\sum_{j \in \mathcal{N}_{i}} \beta_{i j} \boldsymbol{W} \vec{h}_{j}^{l-1}\right),
\end{equation}
\end{small}
where $\vec{h}_i^l$ is the output embedding of $l$-th GAT layer ( $\vec{h}_i^{0} = \vec{h}_i$), $\sigma\left(\cdot\right)$ is a activation function, $ \boldsymbol{W} \in \mathbb{R}^{d\times d}$ is a learnable  shared linear transformation weight matrix. $\mathcal{N}_{i}$ is the neighborhood of $v_i$, i.e., $\forall v_j \in \mathcal{N}_{i} \text{ s.t. } \textbf{A}_{ij} = 1$, and  $\beta_{i j}$ is the importance of $v_j$ to $v_i$:
\begin{small}
\begin{equation}
\beta_{i j}=\frac{\exp \left(\operatorname{LeakyReLU}\left(\vec{\mathbf{a}}^{T}\left[\boldsymbol{W} \vec{h}_{i} \| \boldsymbol{W} \vec{h}_{j}\right]\right)\right)}{\sum_{k \in \mathcal{N}_{i}} \exp \left(\operatorname{LeakyReLU}\left(\vec{\mathbf{a}}^{T}\left[\boldsymbol{W} \vec{h}_{i} \| \boldsymbol{W} \vec{h}_{k}\right]\right)\right)},
\end{equation}
\end{small}
where $\operatorname{LeakyReLU}\left(\cdot\right)$ is one kind of activation function, $\vec{\mathbf{a}}\in \mathbb{R}^{2\cdot d}$ is a learnable shared attention parameter and $\|$ is concatenation operation.

\subsection{Attention-based Sequential Module}
Inspired by the self-attention mechanism, which shows excellent performance in sequential information extraction, we propose our Attention-based Sequential Module~(ASM), which consists of two parts. First, KAAE is proposed. Then, as previous work \cite{lian2020geography, luo2021stan, wang2022spatial} demonstrated, using only a self-attention encoder performs less. We introduce STAAD to model the user preferences for each POI.
\subsubsection{Knowledge-aware Self-Attention Encoder \label{kaae}}
Initially, the knowledge-embedded representations of the user, time nodes, POI, and location nodes, represented by $\vec{h}^2_{u_i},\ \vec{h}^2_{p_j},\ \vec{h}^2_{t_k^d}, \ \vec{h}^2_{t_k^m},\ \vec{h}^2_{t_k^y},\ \vec{h}^2_{g_j^4},\ \vec{h}^2_{g_j^5}, $ and $\vec{h}^2_{g_j^6}$ respectively, are obtained from the HGNN. These embeddings are then concatenated to derive the representation $\vec{c_i}$ of each check-in.

To further encode the check-in and transform it into the same vector space as other entities, a single-layer feed-forward network (SFFN) is employed to obtain the transformed representation $\vec{c_i}'$.
The check-in sequence representation $\boldsymbol{C} = [\vec{c_1}' \| \vec{c_2}' \| \dots \| \vec{c}'_{M}]$  can then be obtained, where the length of each sequence is fixed to $M$. Sequences shorter than the fixed length will be padded, while longer sequences will be truncated.\par
Then, we utilize multi-head attention to capture different level sequential information, which first converts the representation of the check-in sequence $\boldsymbol{C}$ into query, key, and value through three distinct matrices $\boldsymbol{W}_Q,\boldsymbol{W}_K, \boldsymbol{W}_V \in \mathbb{R}^{d\times d}$,
\begin{small}
\begin{equation}
\boldsymbol{Q} = \boldsymbol{C}\boldsymbol{W}_Q,  \boldsymbol{K} = \boldsymbol{C}\boldsymbol{W}_K,  \boldsymbol{V} = \boldsymbol{C}\boldsymbol{W}_V, 
\end{equation}
\end{small}
where $\boldsymbol{Q}, \boldsymbol{K}, \boldsymbol{V} \in \mathbb{R}^{M \times d}$ are query, key and value, respectively.
Then, the sequential information is extracted by each attention head with scaled dot-product attention. And for each head, the dimensions are reduced to $\mathbb{R}^{M\times d_k}$, where $d_k=d/N_h$, to keep the output dimension unchanged and $N_h$ is the number of heads, i.e.
\begin{small}
\begin{equation}
\operatorname{head}_i = \operatorname{Attention}\left(\boldsymbol{Q}\boldsymbol{W}_Q^i,\boldsymbol{K}\boldsymbol{W}_K^i,\boldsymbol{V}{\boldsymbol{W}}_V^i\right),
\end{equation}
\end{small}
where  $\boldsymbol{W}_Q^i, \boldsymbol{W}_K^i, \boldsymbol{W}_V^i \in \mathbb{R}^{d\times d_k}$ is the transformation matrix of the query, key, and value for head $i$, respectively. And 
\begin{small}
\begin{equation}
\operatorname{Attention}\left(\boldsymbol{Q},\boldsymbol{K},\boldsymbol{V}\right)= \operatorname{Softmax}\left(\frac{\boldsymbol{Q}\boldsymbol{K}^T}{\sqrt{d}}\right)\boldsymbol{V}
\end{equation}
\end{small}
In the end, the overall sequential information is obtained by combining the sequential information extracted from each head,
\begin{small}
\begin{equation}
    \boldsymbol{C}' = \operatorname{MultiHead}\left(\boldsymbol{Q}, \boldsymbol{K}, \boldsymbol{V}\right)=\left(\| \text{head}_i\right)\boldsymbol{W}_O,
\end{equation}
\end{small}
where $\boldsymbol{W}_O \in \mathbb{R}^{d\times d}$ is the transformation matrix of the output and $\boldsymbol{C}'\in \mathbb{R}^{M \times d}$ is the attentive result.\par
In the end, we leverage an SFFN to endow the attentive result $\boldsymbol{F}$ with non-linearity.
\subsubsection{Spatial-Temporal-aware Attention Decoder \label{staad}}
To let STAAD be spatial and temporal aware, the embedded POI set $\boldsymbol{P}=[\vec{h}^2_{p_1}\| \dots\| \vec{h}^2_{p_{|P|}}]$ is first obtained from HGNN, then the representation of each POI $\vec{h}^2_{p_i}$ is concatenated with the next check-in time embedding $\boldsymbol{t}_{k}=\left(\vec{h}^2_{t^d_{k}}, \vec{h}^2_{t^m_{k}}, \vec{h}^2_{t^y_{k}}\right)$, and an SFFN is then employed to encode the temporal information to get the final representation of each POI $\vec{h}'_{p_i}$. Then, we use the output of the KAAE $\boldsymbol{F}$ as the key and value. And time-embedded POI set $\boldsymbol{P}'= [\vec{h}'_{p_1} \| \dots \| \vec{h}'_{p_{|P|}}]$ as query to get the user preference for each POI,
\begin{small}
\begin{equation}
    \boldsymbol{S} = \operatorname{MultiHead}\left(\boldsymbol{P}', \boldsymbol{F}, \boldsymbol{F}\right),
\end{equation}
\end{small}
where $\boldsymbol{S} \in \mathbb{R}^{|P|\times d}$ is the user's preference over all POI candidates. And the aforementioned mask is still required in STAAD.
\subsection{Prediction Layer \label{predict}}
The output $\boldsymbol{S}$ of STAAD is the preference over all candidate POIs of user $u_i$ at time $t_{k}$. Then, we calculate the matching score over every POI $p_j$ by inner product,
\begin{equation}
    \hat{y}_{i,j} = \boldsymbol{S}_j \cdot \vec{h}'_{p_j},
\end{equation}
where $\boldsymbol{S}_j$ is the j-th row of $\boldsymbol{S}$, representing the preference of user $u_i$ for POI $p_j$. And the POI with the highest matching score is the recommended next POI.\par
We use cross-entropy as our loss function to optimize the model,
\begin{small}
\begin{equation}
    \mathcal{L}_{POI} = =-\frac{1}{m} \sum_{i=1}^{m} \sum_{j=1}^{n} y_{i, j} \log \hat{y}_{i, j} + \lambda \|\theta\|^2,
\end{equation}
\end{small}
where $y_{i, j}=1$ if user $u_i$ visited $p_j$ at time $t_{k}$ and $y_{i, j}=0$ otherwise. $ \|\theta\|^2$ is the regularization term to avoid over-fitting and $\lambda$ controls the strength of the regularization.

\section{Experiments}
In this section, we design various experiments to demonstrate the effectiveness of our approach. We first introduce the used datasets, evaluation metrics, and baseline methods. Then, we present our empirical results and analysis. We also conduct the ablation study to show the effectiveness of each model part and the hyper-parameter study to test the hyper-parameter sensitivity.

\begin{table}[!t]
\caption{Statistics of the LBSN datasets.}\label{tab:Statistics of the LBSNs datasets}
\centering
\renewcommand\arraystretch{1.2}
% \tiny
% \scriptsize
\footnotesize
\resizebox{\linewidth}{!}{
\begin{tabular}{l|r|r|r|r}
\hline
\textbf{Dataset}  & \textbf{NYC} &\textbf{SP} &\textbf{JK} &\textbf{KL} \\
\hline
\#User       & 3,754 & 3,811  &6,184  &6,324   \\
\#POI        & 3,626 &6,255  & 8,805  &10,804   \\
\#POI side-information & 3,546 & 6,168  & 7,687 & 9,969   \\
\#Check-ins   & 104,991 &247,683   & 376,076 &524,061    \\
\#Friendships  & 12,098 &16,363   &17,798  &34,537   \\
\hline
\#Entities & 8,750 & 11,933 & 16,682  & 18,448\\
\#Facts & 121,185 & 241,056 & 343,622 & 404,256\\
\hline
\end{tabular}}
\end{table}

\subsection{Datasets}
We evaluate our proposed HKGNN model on  four selected cities: New York City (\textbf{NYC}), Jakarta (\textbf{JK}), Kuala Lumpur (\textbf{KL}), and São Paulo(\textbf{SP}) in publicly available real-world LBSN datasets Foursquare~\cite{yang2019revisiting}, which contain check-in records, friendships, and location of POIs. We also extended the datasets with the POI side information obtained from the Foursquare open API. The details of these datasets are provided in Table \ref{tab:Statistics of the LBSNs datasets}.
\par 
For preprocessing, we divide the check-in records of each user into three portions, using the first 70\% for training, the middle 10\% for validation, and the last 20\% for test. We remove users with less than three check-ins as we cannot test them. We use each check-in record as the label and the previous 100 check-in records as the input trajectory. If fewer than 100 check-in records are available, the sequence will be padded to 100. Additionally, we calculate the occurrence of each POI in user check-ins and then select the least 30\% POIs to establish a new dataset for each city, where all labels are the least visited POIs. This is done to evaluate the model's ability to alleviate data sparsity since all POIs in this dataset suffer from data sparsity.

\begin{table*}[t]
 \caption{Experimental results on next POI recommendation in selected four cities.}
 \label{baseline result}
 \centering
 % \tiny
% \scriptsize
\footnotesize
\resizebox{\textwidth}{!}{
 \begin{tabular}{lccccccccccccccccccccc}\toprule
%  \multirow{}{}{} 
    & \multicolumn{5}{c}{NYC} & \multicolumn{5}{c}{SP} & \multicolumn{5}{c}{JK}& \multicolumn{5}{c}{KL}
    \\\cmidrule(lr){2-6}\cmidrule(lr){7-11}\cmidrule(lr){12-16} \cmidrule(lr){17-21}
             & Acc@1  & Acc@5 & Acc@10 & MRR & AR    & Acc@1  & Acc@5 & Acc@10 & MRR & AR   & Acc@1  & Acc@5 & Acc@10 & MRR & AR 
  & Acc@1  & Acc@5 & Acc@10 & MRR & AR\\\midrule
 
    Deepmove & 0.1516 & 0.2994 & 0.3321 & 0.2172 & 634.5 & 0.1712 & 0.3630 & 0.4110 & 0.2460 & 589.1 & 0.0990 & 0.2581 & 0.3247 & 0.1753 & 781.7 & 0.0872 & 0.1699 & 0.218 & 0.1315 & 1215  \\ 
    Flashback & 0.1577 & 0.3221 & 0.3774 & 0.2333 & 651.8 & 0.1828 & 0.3871 & 0.4498 & 0.2765 & 668.4 & 0.1458 & 0.3181 & 0.3860 & 0.2281 & 765.9 & 0.1093 & 0.2435 & 0.3007 & 0.1753 & 1010  \\ 
    LSTPM & 0.1623 & 0.2927 & 0.3351 & 0.2298 & 589.5 & 0.1823 & 0.3437 & 0.3959 & 0.2571 & 533.8 & 0.0907 & 0.1903 & 0.2418 & 0.1434 & 887.1 & 0.0744 & 0.1612 & 0.2070 & 0.1209 & 1129  \\ 
    STAN & 0.1519 & 0.2944 & 0.3488 & 0.2198 & 738.1 & 0.1835 & 0.4007 & 0.4725 & 0.2809 & 497.4 & 0.1384 & 0.3361 & 0.3954 & 0.2253 & 691.1 & 0.1139 & 0.2597 & 0.3310 & 0.1855 & 1023\\ 
    ASGNN & 0.0645 & 0.1439 & 0.1770 & 0.1034 & 1115 & 0.1156 & 0.2365 & 0.2900 & 0.1744 & 965.3 & 0.1012 & 0.2288 & 0.2991 & 0.1654 & 944.5 & 0.0673 & 0.1914 & 0.2572 & 0.1302 & 1065  \\ 
    GraphFlashback  & \underline{0.1776} & \underline{0.3647} & \underline{0.4388} & \underline{0.2647} & \underline{530.4} & \underline{0.1965} & \underline{0.4445} & \underline{0.5297} & \underline{0.3086} & \underline{518.0} & \underline{0.1500} & \underline{0.3619} & \underline{0.4468} & \underline{0.2494} & \underline{576.7} & \underline{0.1176} & \underline{0.2796} & \underline{0.3528} & \underline{0.1972} & \underline{793.6} \\ 
    \midrule
    \textbf{HKGNN} &  \textbf{0.2097} & \textbf{0.3982} & \textbf{0.4651} & \textbf{0.2968} & \textbf{374.1} & \textbf{0.3117} & \textbf{0.5335} & \textbf{0.6031} & \textbf{0.4144} & \textbf{229.4} & \textbf{0.2279} & \textbf{0.4232} & \textbf{0.4947} & \textbf{0.3204} & \textbf{376.9} & \textbf{0.1600} & \textbf{0.3260} & \textbf{0.3924} & \textbf{0.2476} & \textbf{543.3}\\ 
     Improvement & 18.07\% & 9.19\% & 5.99\% & 12.13\% & 29.47\% & 58.63\% & 20.02\% & 13.86\% & 34.28\% & 55.71\% & 51.93\% & 16.94\% & 10.72\% & 28.47\% & 34.65\% & 36.05\% & 16.6\% & 11.22\% & 25.56\% & 31.54\% \\
\bottomrule
\\

 \end{tabular}}

\resizebox{\textwidth}{!}{
 \begin{tabular}{lccccccccccccccccccccc}\toprule
%  \multirow{}{}{} 
    & \multicolumn{5}{c}{NYC\_{least}} & \multicolumn{5}{c}{SP\_{least}} & \multicolumn{5}{c}{JK\_{least}}& \multicolumn{5}{c}{KL\_{least}}
    \\\cmidrule(lr){2-6}\cmidrule(lr){7-11}\cmidrule(lr){12-16} \cmidrule(lr){17-21}
             & Acc@1  & Acc@5 & Acc@10 & MRR & AR    & Acc@1  & Acc@5 & Acc@10 & MRR & AR   & Acc@1  & Acc@5 & Acc@10 & MRR & AR 
  & Acc@1  & Acc@5 & Acc@10 & MRR & AR\\\midrule
 
    Deepmove & 0.0254 & 0.0424 & 0.0508 & 0.0343 & \underline{1345} & 0.0435 & 0.1130 & 0.1391 & 0.0759 & 1739 & 0.0161 & 0.0484 & 0.0645 & 0.0295 & 2419 & 0.0096 & 0.0288 & 0.0481 & 0.0272 & 2801 \\ 
    Flashback & 0.0111 & 0.0660 & 0.0943 & 0.0366 & 1638 & 0.0104 & 0.1188 & 0.1824 & 0.0613 & 1844 & 0.0260 & 0.1295 & 0.1814 & 0.0753 & 2341 & 0.0190 & 0.0900 & 0.1226 & 0.0532 & 3006 \\
    LSTPM & 0.0028 & 0.0464 & 0.0666 & 0.0234 & 1593 & 0.0221 & 0.1084 & 0.1434 & 0.0623 & 1934 & 0.0021 & 0.0482 & 0.0693 & 0.0245 & 3119 & 0.0028 & 0.0295 & 0.0414 & 0.0164 & 4068\\  
    STAN & 0.0175 & 0.0791 & 0.1041 & 0.0474 & 1542 & \underline{0.0717} & \underline{0.2620} & \underline{0.3225} & \underline{0.1514} & \underline{1207} & 0.0524 & \underline{0.1821} & 0.2234 & 0.1097 & 2265 & 0.0327
    & 0.1234 & 0.1643 & 0.0793 & 3115\\  
    ASGNN & 0.0090 & 0.0541 & 0.0721 & 0.0301 & 2105 & 0.0181 & 0.1145 & 0.1506 & 0.0609 & 2278 & 0.0117 & 0.1128 & 0.1440 & 0.0508 & 3180 & 0.0309 & 0.0656 & 0.0734 & 0.0443 & 4180  \\
    GraphFlashback  & \underline{0.0315} & \underline{0.1191} & \underline{0.1522} & \underline{0.0739} & 1484 & 0.0454 & 0.2005 & 0.2880 & 0.1177 & 1692 & \underline{0.0557} & 0.1817 & \underline{0.2417} & \underline{0.1170} & \underline{1982} & \underline{0.0506} & \underline{0.1481} & \underline{0.1949} & \underline{0.0979} & \underline{2713}  \\ 
    \midrule
    \textbf{HKGNN}  & \textbf{0.0630} & \textbf{0.1736} & \textbf{0.2144} & \textbf{0.1144} & \textbf{966.9} & \textbf{0.1600} & \textbf{0.3678} & \textbf{0.4362} & \textbf{0.2538} & \textbf{766.3} & \textbf{0.1397} & \textbf{0.3048} & \textbf{0.3463} & \textbf{0.2135} & \textbf{1197} & \textbf{0.0990} & \textbf{0.2230} & \textbf{0.2605} & \textbf{0.1559} & \textbf{1983}\\

    Improvement & 100.0\% & 45.76\% & 40.87\% & 54.8\% & 34.87\% & 123.15\% & 40.38\% & 35.25\% & 67.63\% & 36.51\% & 150.81\% & 67.38\% & 43.28\% & 82.48\% & 39.61\% & 95.65\% & 50.57\% & 33.66\% & 59.24\% & 26.91\% \\

\bottomrule
 \end{tabular}}
 
\end{table*}

\subsection{Baseline Models}
We compare our HKGNN with the following models:
\begin{itemize}
    \item \textbf{DeepMove}\cite{deepmove}: This method incorporates attention mechanism in recurrent neural network to model multi-level periodicity and short-term preferences.
    \item \textbf{Flashback}\cite{flashback}: A RNN-based model that flashback on hidden states to better model sparse user mobility traces.
    \item \textbf{LSTPM}\cite{lstpm}: This method is based on LSTM network and uses the captured long-, short-term user preferences to make the recommendation.
    \item \textbf{STAN}\cite{luo2021stan}: This method uses a self-attention mechanism to explicitly exploit relative spatiotemporal information of all check-ins in the trajectory.
    \item \textbf{ASGNN}\cite{asgnn}: A GNN-based approach, which model the user check-ins as a graph, then uses GAT to capture the user's long- and short-term preferences.  
    \item \textbf{GraphFlashback}\cite{rao2022graph}: A state-of-the-art method that constructs a KG based on the check-ins. Then, Graph Convolution Network is applied to the learned POI transition graph to refine the embedding and employ an RNN-based network to capture the sequential transition patterns.
    
\end{itemize}

\subsection{Evaluation Metrics}
We use three widely used evaluation metrics in recommendation: Accuracy@K~(Acc@K), Mean Reciprocal Rank~ (MRR), and Average Rank~(AR) to evaluate the performance of HKGNN. AR is used to evaluate the performance of the model on difficult samples since Acc@K and MRR are not sensitive to those low-rank predictions.

\subsection{Experimental Settings}
 For all baseline models, we follow the recommended experimental settings from the original paper and keep the hyperparameters unchanged.  For STAN, we do not follow the instruction from its released code that samples a small proportion of users to train the model each time since  the training time is acceptable.\par
 For our HKGNN, we use ReLU as the activation function $\sigma(\cdot)$. And we fine-tune the hyperparameter settings through the validation set. The embedding dimension $d$ is set to 256, the attention heads $N_h$ in KAAE and STAAD are set to 4, and the negative sampling rate $N$ in HKG training is set to 10. We train our model with Adam optimizer and a learning rate of 0.01 for HKG training and $10^{-5}$ for HGNN and ASM training. The dropout rate and regularization term $\lambda$ is set to 0.2 and 0.0001, respectively, to reduce over-fitting. We train 100 epochs for HKG and 50 epochs for HGNN and ASM. For the larger two datasets, JK and KL, we set the learning rate and regularization term to $4\cdot10^{-6}$ and $10^{-4}$, respectively, for better convergence. Our implementation is available in Pytorch.

\subsection{Experimental Results and Analysis}
The experimental results for four selected datasets along with their corresponding least visited datasets are shown in Table \ref{baseline result}. Our HKGNN model consistently outperforms all baseline models across all metrics on these datasets. Notably, on the NYC dataset, HKGNN demonstrates significant improvements in Acc@1, Acc@5, Acc@10, MRR, and AR by 18.07\%, 9.19\%, 5.99\%, 12.13\%, and 29.47\%, respectively, which demonstrates the efficacy of our approach.

In the case of datasets containing the least visited POIs, all models exhibit varying degrees of performance degradation due to data sparsity. Sequential-based methods like DeepMove, LSTPM, and Flashback struggle to capture user preferences for these POIs, given their infrequent occurrences in training data. Graph-FlashBack, among the baseline models, performs comparatively better on these datasets, highlighting the benefit of knowledge graphs in mitigating data sparsity. Our HKGNN, on the other hand, leveraging hyper-relations in LBSN and combining semantic and structural information from HKG, further enhances performance on these least visited POIs.

Unlike Graph-FlashBack, which employs a KG and RNN architecture, our method utilizes a HKG and self-attention network. Our hyper-relational design and the application of HGNN sets us apart, as it enables us to leverage the various hyper-relations in LBSN and the structural infomation.

HKGNN exhibits relatively larger improvements in Acc@1 and AR metrics, underscoring its ability to provide high-quality recommendations for all POIs and address the data sparsity challenge, resulting in substantial enhancements for recommendations to low-rank POIs.

\begin{table}[t]
 \caption{Ablation study results. w/o \textit{component-name} is the variant that removes a certain component.}
 \label{ablation result}
 \centering
% \scriptsize
\footnotesize
\resizebox{\linewidth}{!}{
 \begin{tabular}{lcccccccc}\toprule
%  \multirow{}{}{} 
     & \multicolumn{4}{c}{SP} & \multicolumn{4}{c}{SP\_{least}}
    \\\cmidrule(lr){2-5}\cmidrule(lr){6-9}
             & Acc@1  & Acc@5 & Acc@10  & AR & Acc@1  & Acc@5 & Acc@10 & AR\\\midrule
 
    \textbf{HKGNN}  & \textbf{0.3117} & \textbf{0.5335} & 0.6031 & \textbf{229.4}  & \textbf{0.1600} & \textbf{0.3678} & \textbf{0.4362} & \textbf{766.3}\\
    w/o HKG  & 0.3051 & 0.5231 & 0.5945 & 233.5 & 0.1194 & 0.3213 & 0.3951 & 783.8 \\
    w/o HGNN & 0.2905 & 0.5321 & \textbf{0.6079} & 474.9 & 0.0977 & 0.3082 & 0.3884 & 1114\\
    w/o side-info & 0.2531 & 0.4741 & 0.5448 & 282.7 & 0.0921 & 0.2848 & 0.3524 & 969.5 \\
    w/o social & 0.2584 & 0.4704 & 0.5442 & 336.0 & 0.1083 & 0.2883 & 0.3518 & 1142.4 \\
    w/o mobility & 0.2940 & 0.5311 & 0.6072 & 429.3 & 0.1240 & 0.3369 & 0.4229 & 1216 \\

\bottomrule
 \end{tabular}}
\end{table}

\subsection{Ablation Study}
In this section, we conduct several ablation experiments to demonstrate the effectiveness of each model part. Specifically, we design two variants that remove the HKG and HGNN separately. In addition, to evaluate the impact of the proposed three types of relations, we also remove them individually.
We test these variants on \textbf{SP} dataset, and the results are shown in Table \ref{ablation result}. Based on these results, we have the following analysis:\looseness=-1\par
$\bullet$ Both HKG and HGNN contribute to the model performance, demonstrating their effectiveness. When removing HKG and HGNN from the model, the performance decreased more on the least visited dataset, which shows the strong ability of our HKG to tackle data sparsity and the importance of structural information in HKG, due to the higher-order relations of entities it contains.\par
$\bullet$ All three relations play vital roles in POI recommendation. In particular, social and side-info relations are most significant, while mobility pattern relation affects less. In addition, these relations are all essential for those least visited POI, which shows their ability to alleviate the data sparsity.

\begin{figure}[htbp]
  \centering
  
  \begin{subfigure}[b]{0.49\linewidth}
    \centering
    \includegraphics[width=\linewidth]{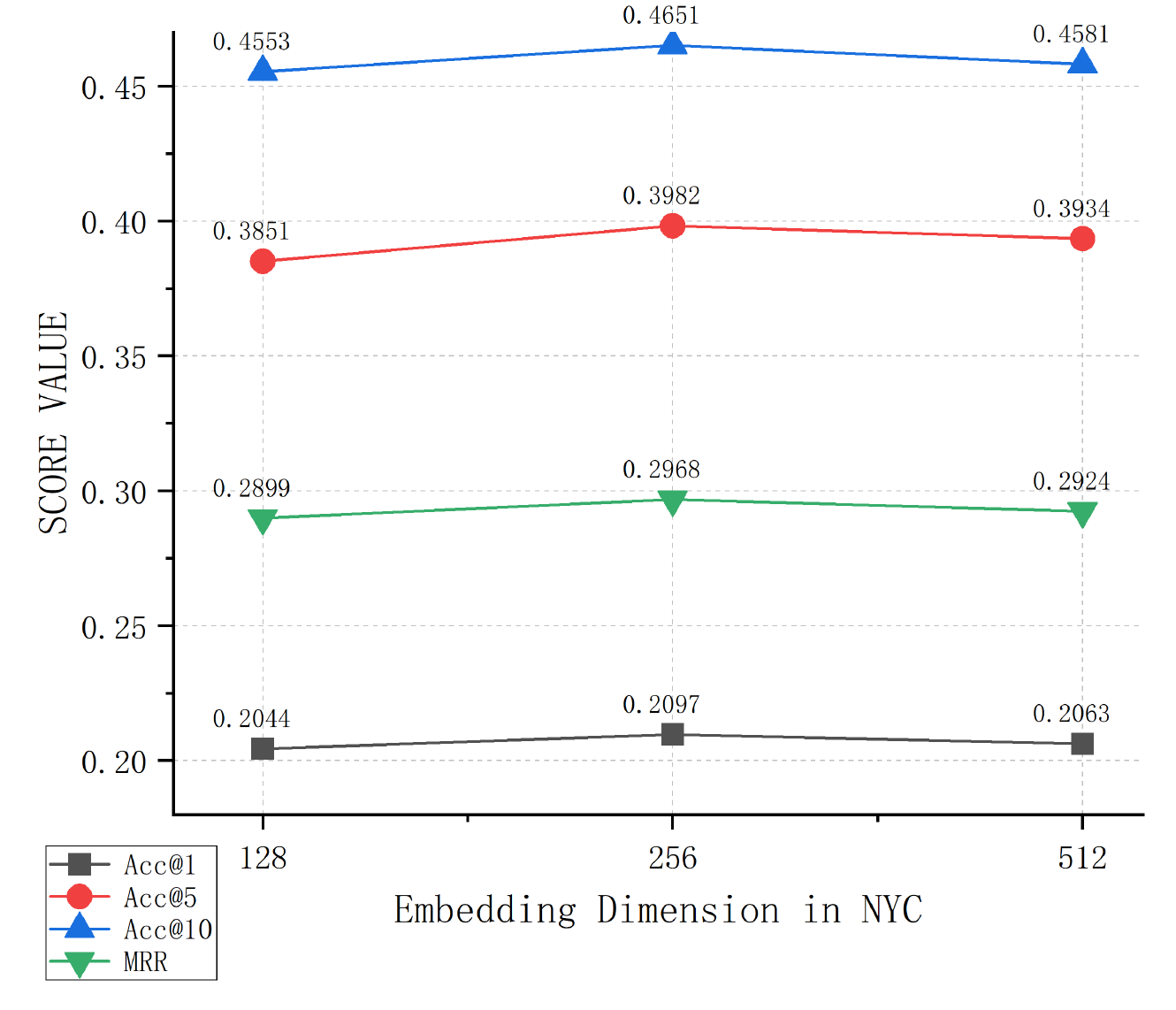}
    \caption{Impact of dimension $d$}
    \label{fig:result_nyc}
  \end{subfigure}
  % \hfill
  \centering
  \begin{subfigure}[b]{0.49\linewidth}
    \centering
    \includegraphics[width=\linewidth]{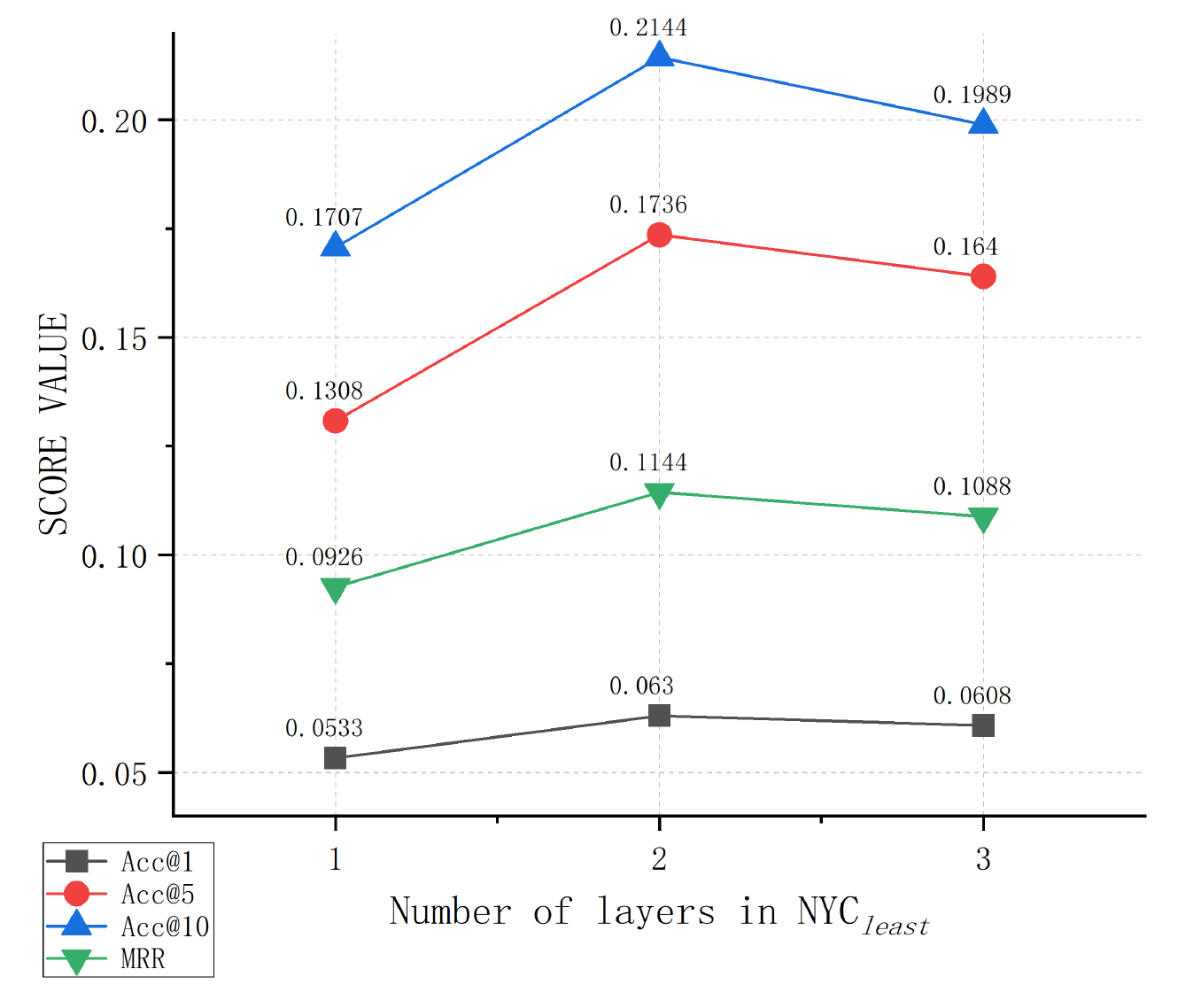}
    \caption{Impact of GAT layers}
    \label{fig:impact_gat_layer}
  \end{subfigure}
  
  \caption{Result of hyperparameter study}
  \label{fig:hyperparameter_test}
\end{figure}
\subsection{Hyperparameter Study}
This section aims to provide a comprehensive analysis of the influence of different hyperparameter settings on the proposed HKGNN model. We systematically investigate the impact of two key hyperparameters: the embedding dimension~($d$) and the number of layers in the HGNN.   \par
The results are depicted in Fig.~\ref{fig:hyperparameter_test}. The model's performance is influenced by embedding dimensions, with both lower and higher dimensions having a negative impact~(Fig.~\ref{fig:result_nyc}). This suggests that a small number of features fails to fully capture the semantics of HKG, while larger dimensions could lead to convergence issues due to redundant features. Thus, we set the embedding dimension as $d=256$. Furthermore, performance degradation occurs when the layer count exceeds 2~(Fig.~\ref{fig:impact_gat_layer}), this is possibly attributed to the over-smoothing problem in GNNs.\looseness=-1
 \section{Conclusion}
 In this paper, we present a novel Hyper-relational Knowledge Graph Neural Network model for the next POI recommendation. We use an HKG to model the hyper-relations in LBSN and utilize an HGNN to leverage the structural information in HKG. We also exploit untapped side information in LBSN. Experimental results on four real-world datasets demonstrate the effectiveness of our proposed model in alleviating data sparsity. We also conduct an ablation study, which confirms the effectiveness of each model part.

\bibliography{aaai24}

\begin{thebibliography}{31}
\providecommand{\natexlab}[1]{#1}

\bibitem[{Agarwal, Branson, and Belongie(2006)}]{star}
Agarwal, S.; Branson, K.; and Belongie, S. 2006.
\newblock Higher order learning with graphs.
\newblock In \emph{Proceedings of the 23rd international conference on Machine learning}, 17--24.

\bibitem[{Bagci and Karagoz(2016)}]{friend1}
Bagci, H.; and Karagoz, P. 2016.
\newblock Context-aware friend recommendation for location based social networks using random walk.
\newblock In \emph{Proceedings of the 25th international conference companion on world wide web}, 531--536.

\bibitem[{Chen et~al.(2022)Chen, Wan, Guo, Huang, Zheng, Li, Lin, and Lin}]{chen2022building}
Chen, W.; Wan, H.; Guo, S.; Huang, H.; Zheng, S.; Li, J.; Lin, S.; and Lin, Y. 2022.
\newblock Building and exploiting spatial--temporal knowledge graph for next POI recommendation.
\newblock \emph{Knowledge-Based Systems}, 258: 109951.

\bibitem[{Cui et~al.(2021)Cui, Sun, Zhao, Yin, and Zheng}]{meta2021}
Cui, Y.; Sun, H.; Zhao, Y.; Yin, H.; and Zheng, K. 2021.
\newblock Sequential-knowledge-aware next POI recommendation: A meta-learning approach.
\newblock \emph{ACM Transactions on Information Systems (TOIS)}, 40(2): 1--22.

\bibitem[{Fatemi et~al.(2019)Fatemi, Taslakian, Vazquez, and Poole}]{hsimple}
Fatemi, B.; Taslakian, P.; Vazquez, D.; and Poole, D. 2019.
\newblock Knowledge hypergraphs: Prediction beyond binary relations.
\newblock \emph{arXiv preprint arXiv:1906.00137}.

\bibitem[{Feng et~al.(2018)Feng, Li, Zhang, Sun, Meng, Guo, and Jin}]{deepmove}
Feng, J.; Li, Y.; Zhang, C.; Sun, F.; Meng, F.; Guo, A.; and Jin, D. 2018.
\newblock Deepmove: Predicting human mobility with attentional recurrent networks.
\newblock In \emph{Proceedings of the 2018 world wide web conference}, 1459--1468.

\bibitem[{Feng et~al.(2015)Feng, Li, Zeng, Cong, and Chee}]{feng2015personalized}
Feng, S.; Li, X.; Zeng, Y.; Cong, G.; and Chee, Y.~M. 2015.
\newblock Personalized ranking metric embedding for next new poi recommendation.
\newblock In \emph{IJCAI'15 Proceedings of the 24th International Conference on Artificial Intelligence}, 2069--2075. ACM.

\bibitem[{Guo et~al.(2020)Guo, Sun, Zhang, and Theng}]{guo2020attentional}
Guo, Q.; Sun, Z.; Zhang, J.; and Theng, Y.-L. 2020.
\newblock An attentional recurrent neural network for personalized next location recommendation.
\newblock In \emph{Proceedings of the AAAI Conference on artificial intelligence}, volume~34, 83--90.

\bibitem[{Huang et~al.(2022)Huang, Ma, Dong, Foutz, and Li}]{poi2}
Huang, Z.; Ma, J.; Dong, Y.; Foutz, N.~Z.; and Li, J. 2022.
\newblock Empowering Next POI Recommendation with Multi-Relational Modeling.
\newblock In \emph{Proceedings of the 45th International ACM SIGIR Conference on Research and Development in Information Retrieval}, 2034--2038.

\bibitem[{Koren(2009)}]{koren2009collaborative}
Koren, Y. 2009.
\newblock Collaborative filtering with temporal dynamics.
\newblock In \emph{Proceedings of the 15th ACM SIGKDD international conference on Knowledge discovery and data mining}, 447--456.

\bibitem[{Li, Shen, and Zhu(2018)}]{li2018next}
Li, R.; Shen, Y.; and Zhu, Y. 2018.
\newblock Next point-of-interest recommendation with temporal and multi-level context attention.
\newblock In \emph{2018 IEEE International Conference on Data Mining (ICDM)}, 1110--1115. IEEE.

\bibitem[{Li et~al.(2021)Li, Chen, Luo, Yin, and Huang}]{ijcai2021p206}
Li, Y.; Chen, T.; Luo, Y.; Yin, H.; and Huang, Z. 2021.
\newblock Discovering Collaborative Signals for Next POI Recommendation with Iterative Seq2Graph Augmentation.
\newblock In Zhou, Z.-H., ed., \emph{Proceedings of the Thirtieth International Joint Conference on Artificial Intelligence, {IJCAI-21}}, 1491--1497. International Joint Conferences on Artificial Intelligence Organization.
\newblock Main Track.

\bibitem[{Li et~al.(2022)Li, Fan, Zhang, Shi, Xu, Yin, Deng, and Song}]{friend3}
Li, Y.; Fan, Z.; Zhang, J.; Shi, D.; Xu, T.; Yin, D.; Deng, J.; and Song, X. 2022.
\newblock Heterogeneous Hypergraph Neural Network for Friend Recommendation with Human Mobility.
\newblock In \emph{Proceedings of the 31st ACM International Conference on Information \& Knowledge Management}, 4209--4213.

\bibitem[{Lian et~al.(2020)Lian, Wu, Ge, Xie, and Chen}]{lian2020geography}
Lian, D.; Wu, Y.; Ge, Y.; Xie, X.; and Chen, E. 2020.
\newblock Geography-aware sequential location recommendation.
\newblock In \emph{Proceedings of the 26th ACM SIGKDD international conference on knowledge discovery \& data mining}, 2009--2019.

\bibitem[{Lim et~al.(2022)Lim, Hooi, Ng, Goh, Weng, and Tan}]{lim2022hierarchical}
Lim, N.; Hooi, B.; Ng, S.-K.; Goh, Y.~L.; Weng, R.; and Tan, R. 2022.
\newblock Hierarchical multi-task graph recurrent network for next poi recommendation.
\newblock In \emph{Proceedings of the 45th international ACM SIGIR conference on Research and development in Information Retrieval}.

\bibitem[{Lim et~al.(2020)Lim, Hooi, Ng, Wang, Goh, Weng, and Varadarajan}]{lim2020stp}
Lim, N.; Hooi, B.; Ng, S.-K.; Wang, X.; Goh, Y.~L.; Weng, R.; and Varadarajan, J. 2020.
\newblock STP-UDGAT: Spatial-temporal-preference user dimensional graph attention network for next POI recommendation.
\newblock In \emph{Proceedings of the 29th ACM International Conference on Information \& Knowledge Management}, 845--854.

\bibitem[{Luo, Liu, and Liu(2021)}]{luo2021stan}
Luo, Y.; Liu, Q.; and Liu, Z. 2021.
\newblock Stan: Spatio-temporal attention network for next location recommendation.
\newblock In \emph{Proceedings of the Web Conference 2021}, 2177--2185.

\bibitem[{Qian et~al.(2019)Qian, Liu, Nguyen, and Yin}]{qian2019spatiotemporal}
Qian, T.; Liu, B.; Nguyen, Q. V.~H.; and Yin, H. 2019.
\newblock Spatiotemporal representation learning for translation-based POI recommendation.
\newblock \emph{ACM Transactions on Information Systems (TOIS)}, 37(2): 1--24.

\bibitem[{Rao et~al.(2022)Rao, Chen, Liu, Shang, Yao, and Han}]{rao2022graph}
Rao, X.; Chen, L.; Liu, Y.; Shang, S.; Yao, B.; and Han, P. 2022.
\newblock Graph-flashback network for next location recommendation.
\newblock In \emph{Proceedings of the 28th ACM SIGKDD Conference on Knowledge Discovery and Data Mining}, 1463--1471.

\bibitem[{Sun et~al.(2020)Sun, Qian, Chen, Liang, Nguyen, and Yin}]{lstpm}
Sun, K.; Qian, T.; Chen, T.; Liang, Y.; Nguyen, Q. V.~H.; and Yin, H. 2020.
\newblock Where to go next: Modeling long-and short-term user preferences for point-of-interest recommendation.
\newblock In \emph{Proceedings of the AAAI Conference on Artificial Intelligence}, volume~34, 214--221.

\bibitem[{Veli{\v{c}}kovi{\'c} et~al.(2018)Veli{\v{c}}kovi{\'c}, Cucurull, Casanova, Romero, Li{\`o}, and Bengio}]{gat2018}
Veli{\v{c}}kovi{\'c}, P.; Cucurull, G.; Casanova, A.; Romero, A.; Li{\`o}, P.; and Bengio, Y. 2018.
\newblock Graph Attention Networks.
\newblock In \emph{International Conference on Learning Representations}.

\bibitem[{Wang et~al.(2021{\natexlab{a}})Wang, Wang, Xiang, Yu, Deng, and Xu}]{asgnn}
Wang, D.; Wang, X.; Xiang, Z.; Yu, D.; Deng, S.; and Xu, G. 2021{\natexlab{a}}.
\newblock Attentive sequential model based on graph neural network for next poi recommendation.
\newblock \emph{World Wide Web}, 24(6): 2161--2184.

\bibitem[{Wang et~al.(2022)Wang, Jiang, Xu, Wang, and Yang}]{wang2022spatial}
Wang, E.; Jiang, Y.; Xu, Y.; Wang, L.; and Yang, Y. 2022.
\newblock Spatial-Temporal Interval Aware Sequential POI Recommendation.
\newblock In \emph{2022 IEEE 38th International Conference on Data Engineering (ICDE)}, 2086--2098. IEEE.

\bibitem[{Wang et~al.(2021{\natexlab{b}})Wang, Yu, Liu, Jin, and Li}]{wang2021spatio}
Wang, H.; Yu, Q.; Liu, Y.; Jin, D.; and Li, Y. 2021{\natexlab{b}}.
\newblock Spatio-temporal urban knowledge graph enabled mobility prediction.
\newblock \emph{Proceedings of the ACM on interactive, mobile, wearable and ubiquitous technologies}, 5(4): 1--24.

\bibitem[{Wang et~al.(2018)Wang, Zhang, Xie, and Guo}]{wang2018dkn}
Wang, H.; Zhang, F.; Xie, X.; and Guo, M. 2018.
\newblock DKN: Deep knowledge-aware network for news recommendation.
\newblock In \emph{Proceedings of the 2018 world wide web conference}, 1835--1844.

\bibitem[{Wang et~al.(2019)Wang, He, Cao, Liu, and Chua}]{wang2019kgat}
Wang, X.; He, X.; Cao, Y.; Liu, M.; and Chua, T.-S. 2019.
\newblock Kgat: Knowledge graph attention network for recommendation.
\newblock In \emph{Proceedings of the 25th ACM SIGKDD international conference on knowledge discovery \& data mining}, 950--958.

\bibitem[{Yang et~al.(2020)Yang, Fankhauser, Rosso, and Cudre-Mauroux}]{flashback}
Yang, D.; Fankhauser, B.; Rosso, P.; and Cudre-Mauroux, P. 2020.
\newblock Location prediction over sparse user mobility traces using rnns.
\newblock In \emph{Proceedings of the twenty-ninth international joint conference on artificial intelligence}, 2184--2190.

\bibitem[{Yang et~al.(2019)Yang, Qu, Yang, and Cudre-Mauroux}]{yang2019revisiting}
Yang, D.; Qu, B.; Yang, J.; and Cudre-Mauroux, P. 2019.
\newblock Revisiting user mobility and social relationships in lbsns: a hypergraph embedding approach.
\newblock In \emph{The world wide web conference}, 2147--2157.

\bibitem[{Zhang et~al.(2020)Zhang, Li, Gou, and Yang}]{zhang2020kean}
Zhang, C.; Li, T.; Gou, Y.; and Yang, M. 2020.
\newblock KEAN: Knowledge embedded and attention-based network for POI recommendation.
\newblock In \emph{2020 IEEE International Conference on Artificial Intelligence and Computer Applications (ICAICA)}, 847--852. IEEE.

\bibitem[{Zhang et~al.(2016)Zhang, Yuan, Lian, Xie, and Ma}]{zhang2016collaborative}
Zhang, F.; Yuan, N.~J.; Lian, D.; Xie, X.; and Ma, W.-Y. 2016.
\newblock Collaborative knowledge base embedding for recommender systems.
\newblock In \emph{Proceedings of the 22nd ACM SIGKDD international conference on knowledge discovery and data mining}, 353--362.

\bibitem[{Zhou, Huang, and Sch{\"o}lkopf(2006)}]{ce}
Zhou, D.; Huang, J.; and Sch{\"o}lkopf, B. 2006.
\newblock Learning with hypergraphs: Clustering, classification, and embedding.
\newblock \emph{Advances in neural information processing systems}, 19.

\end{thebibliography}

\end{document}